\documentclass{article}
\usepackage{ijcai21}

\usepackage{times}
\usepackage{soul}
\usepackage{url}
\usepackage[hidelinks]{hyperref}
\usepackage[utf8]{inputenc}
\usepackage[small]{caption}
\usepackage{graphicx}
\usepackage{amssymb,amsmath}
\usepackage{amsthm}
\usepackage{booktabs}
\usepackage{algorithm}
\usepackage{algorithmic}
\title{Memory-Associated Differential Learning}

\author{
Yi Luo$^1$
\and
Aiguo Chen$^1$\footnote{Corresponding Author}\and
Bei Hui$^2$\and
Ke Yan$^1$\\
\affiliations
$^1$School of Computer Science and Engineering, University of Electronic Science and Technology of China, Chengdu 611731, China.\\
$^2$School of Information and Software Engineering, University of Electronic Science and Technology of China, Chengdu 611731, China.\\
\emails
cf020031308@163.com,
\{agchen, bhui, kyan\}@uestc.edu.cn
}

\begin{document}
\maketitle

\begin{abstract}
Conventional Supervised Learning approaches focus on the mapping from
input features to output labels. After training, the learnt models alone
are adapted onto testing features to predict testing labels in isolation,
with training data wasted and their associations ignored. To take full
advantage of the vast number of training data and their associations,
we propose a novel learning paradigm called
\emph{Memory-Associated Differential (MAD) Learning}. We first introduce
an additional component called \emph{Memory} to memorize all the training
data. Then we learn the differences of labels as well as the associations
of features in the combination of a differential equation and some
sampling methods. Finally, in the evaluating phase, we predict unknown
labels by inferencing from the memorized facts plus the learnt
differences and associations in a geometrically meaningfull manner. We
gently build this theory in unary situations and apply it on Image
Recognition, then extend it into Link Prediction as a binary situation,
in which our method outperforms strong state-of-the-art baselines on
ogbl-ddi dataset.
\end{abstract}

\hypertarget{introduction}{%
\section{Introduction}
\label{sec:introduction}}


\begin{figure}
\centering
\includegraphics[bb=0 0 573 253, width=0.5\textwidth]{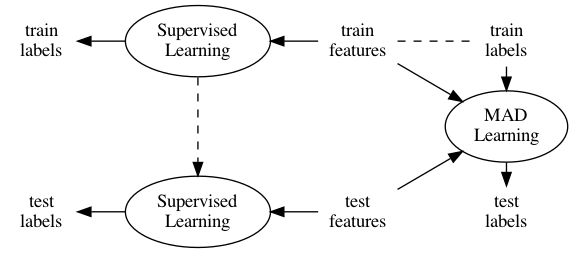}
\caption{
  The difference between Conventional Supervised Learning and MAD Learning.
  The former learns the mapping from features to labels in training data
  and apply this mapping on testing data, while the latter learns the
  differences and associations among data and inferences testing labels
  from memorized training data.
}
\label{fig:sl-mad}
\end{figure}

In this work, we develop \emph{MAD Learning,
Memory-Associated Differential Learning}, to inference from the
memorized facts that we already know to predict what we want to know.
Different from conventional Supervised Learning
approaches which learn the mapping from input features to output labels,
our method focuses on the relationship between features and features,
labels and labels. When predicting on testing features, MAD Learning
inferences from similar memory and make some difference upon it to get
testing labels. We illustrate this difference in Figure~\ref{fig:sl-mad}.

In Section~\ref{ssec:memory-associated-differential-learning}, we gently
build this theory in the unary situation. To illustrate that the existence of
memory and differences are equally important, we conduct
experiments with either or none of these two parts disabled in
Section~\ref{sssec:memory-and-differential-functions}.


To have a better approximation, we investigate several techniques of
sampling such as multi-heads~\cite{1706.03762v5} and propose
a mechanism called \emph{Soft Sentinel} to softly filter out unreliable
estimations. In Section~\ref{sssec:aggregators-and-soft-sentinel} we
examine the effectiveness of these techniques and mechanism.

We then extend MAD Learning into binary situations in
Section~\ref{sssec:link-prediction} where Link
Prediction~\cite{sun2011co} is taken as an example. In
Section~\ref{ssec:link-prediction} we experiment on
dataset \emph{ogbl-ddi}~\cite{wishart2018drugbank}
from Open Graph Benchmark (OGB)~\cite{hu2020ogb}.
On ogbl-ddi, MAD Learning outperforms state-of-the-art (SotA) baselines.

Since the differential equation in MAD Learning has the form of
the first-order Taylor series approximation, it gains clear
interpretability in geometry. In
Section~\ref{sssec:the-geometric-meaning-of-mad-learning}, we visualize
a social network Zachary's Karate Club~\cite{zachary1977information} to
reveal the meaning of learnt encodings.

Finally, we extend MAD Learning to ternary and multi situations to
construct more complex applications such as relation predictors for
Knowledge Graphs. However, due to the huge space occupation of this
method, this extension remains as a theory that we place it into
Section~\ref{sec:discussion} as the discussion.

As a novel learning paradigm, MAD Learning opens the door for many
research directions. We raise some of them in
Section~\ref{sec:conclusion} as conclusions.

\hypertarget{related-works}{%
\section{Related Works}\label{sec:related-works}}

Known facts to Machine Learning models are just memorized experience to
human. Early in 2014, researchers in Natural Language Processing
believed the internal memory a Recurrent Neural
Network~\cite{DBLP:conf/interspeech/MikolovKBCK10} had so insufficient to
accurately remember all the facts occurred in history, that they
proposed Memory Networks~\cite{DBLP:journals/corr/WestonCB14} to take
advantage of historical facts by writing to and reading from External Memory.
A similar idea is adopted by us but in a different way. Instead of
treating External Memory as a way to add more learnable parameters to
store uninterpretable hidden states, we try to memorize the facts as
they are, and then learn the differences and associations between them.

Most of the experiments in this article are designed to solve Link
Prediction problem that we predict whether a pair of nodes in a graph
are likely to be connected, how much the weight their edge bares, or
what attributes their edge should have. In such a field, two of the most
popular methods are Graph Convolution Networks
(GCN)~\cite{DBLP:journals/corr/KipfW16} and Matrix
Factorization~\cite{koren2009matrix}.

The idea of GCN is that the hidden representation of a node can be
aggregated from the states of itself and of its neighbours, this
usually implies that connected nodes should have similar
representations. However, this assumption results in
\emph{over-smoothing}~\cite{DBLP:conf/iccv/Li0TG19} that the
representations of nodes become nearly identical after multiple
layers that they can hardly be distinguished.
MAD Learning on graphs does not suffer from this issue. It has a loose
constraint that the neighbours of the same node ought to have similar
representations while the connected pairs do not have to. This matters
when edges are directional.

Matrix Factorization is a classical algorithm used in Recommender Systems.
It decomposes the adjacency
matrix into the product of two matrices. Each works as a group of embeddings
for nodes. Although our method is derived from a different perspective
of view, we point out that Matrix Factorization can be seen as a
simplification of MAD Learning with no memory and no sampling.

\hypertarget{proposed-approach}{%
\section{Proposed Approach}\label{sec:proposed-approach}}

\hypertarget{memory-associated-differential-learning}{%
\subsection{Memory-Associated Differential Learning}\label{ssec:memory-associated-differential-learning}}

We assume that the output label \(y\) of an instance of input features
\(x\) is a differentiable function \(y = y(x)\).
And besides bare features \(x\) with its label \(y\) unknown, we also
have another \(x_0\) called \emph{reference} with its output label
\(y_0\) already known.

By applying Mean Value Theorem for Definite
Integrals~\cite{comenetz2002calculus}, we can estimate
the unknown \(y\) with known \(y_0\) if \(x_0\) is close enough to \(x\):
\[y = y(x) = y_0 + \int_{x_0}^x dy
\approx y_0 + (x - x_0) \cdot y'(x)
\triangleq \hat y | y_0\]

In such way, we connect the current prediction tasks \(y\) to the past
fact \(y_0\), which can be stored in external memory, and convert the
learning of our target function \(y(x)\) to the learning of a
differential function \(y'(x)\), which in general is more accessible
than the former.

\begin{figure*}
\centering
\includegraphics[bb=0 0 1041 341, width=\textwidth]{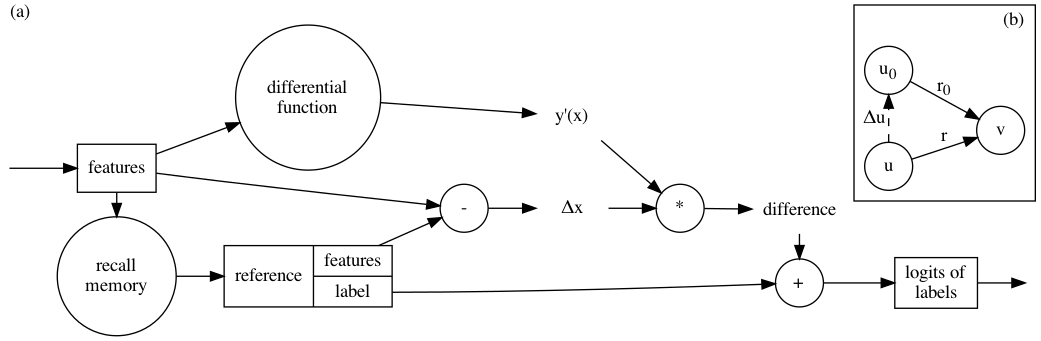}
\caption{
  (a) Memory-Associated Differential Learning
  inferences labels from memorized ones following the first-order Taylor
  series approximation: \( y \approx y_0 + \Delta x \cdot y'(x) \).
  (b) In binary MAD Learning, when \( v = v_0 \) holds,
  \( \frac{\partial r}{\partial u}|_{(u, v)} \) is simplified to be
  \( \frac{\partial r}{\partial u}|_v \) since it is the change of \( r \)
  after slightly moving \( u \) to \( u_0 \) with \( v \) fixed.
}
\label{fig:mad-flow}
\end{figure*}

In Figure~\ref{fig:mad-flow}(a), we depict how MAD Learning predict the
distribution of labels by incorporating memory and the learnable
differential function. The importance of these two parts is verified
in Section~\ref{sssec:memory-and-differential-functions}.

\hypertarget{inferencing-from-multiple-references}{%
\subsection{Inferencing from Multiple References}\label{ssec:inferencing-from-multiple-references}}

To get a steady and accurate estimation of \(y\), we can sample \(n\)
references \(x_1, x_2, \cdots, x_n\) to get \(n\)
estimations \(\hat y | y_1, \hat y | y_2, \cdots, \hat y | y_n\) and
combine them with an aggregator such as \emph{mean}:
\[\hat y = \frac{1}{n} \cdot \sum\limits_{i=1}^{n} \hat y | y_i\]

Since the closer the reference \(x_i\) is, the more accurate the
estimation \(\hat y | y_i\) it gives, we can attach a weight on each
estimation inversely proportional to the distance between the reference
\(x_i\) and \(x\). Here we adopt a function \emph{Softmin} derived from
Softmax which rescales the inputted \(d\)-dimentional array \(v\) so
that every element of \(v\) lies in the range \([0, 1]\) and all of them
sum to 1:
\[\text{Softmin}(v_i) = \text{Softmax}(-v_i) = \frac{e^{-v_i}}{\sum\limits_{j=1}^{d} e^{-v_j}}\]

By applying Softmin we get the aggregated estimation: \[\begin{cases}
\hat y &= \frac{1}{Z} \cdot \sum\limits_{i=1}^{n} \hat y | y_i \cdot e^{-||x_i - x||} \\
Z &= \sum\limits_{i=1}^{n} e^{-||x_i - x||}
\end{cases}\]

\hypertarget{soft-sentinels-and-uncertainty}{%
\subsection{Soft Sentinels and Uncertainty}\label{ssec:soft-sentinels-and-uncertainty}}

With Softmin, inaccurate estimations given by distant
references can hardly distort the final result, if but only if
nearby references exist. Otherwise, a group of distant references which
gives less reliable estimations also has a summed weight of 1,
the same as a group of close references.

To rectify this issue, we introduce a mechanism on top of Softmin named
\emph{Soft Sentinel}. A Soft Sentinel is a dummy element mixed into the
array of estimations with no information (e.g.~the logit is 0) but a set
distance (e.g.~0).

The estimation after \(k\) Soft Sentinels distant at 1 added is \[\begin{cases}
\hat y &= \frac{1}{Z} \cdot \sum\limits_{i=1}^{n} \hat y | y_i \cdot e^{-||x_i - x||}\\
Z &= ke^{-1} + \sum\limits_{i=1}^{n} e^{-||x_i - x||}
\end{cases}\]

When Soft Sentinels involved, only estimations given by close-enough
references can have most of their impacts on the final result that unreliable
estimations are supressed.

Furthermore, The weight of a single Soft Sentinel distant at 0 can be
viewed as a measure of uncertainty when predicting: the further the
references are, the more the uncertainty is.

In Section~\ref{sssec:aggregators-and-soft-sentinel}, along with the
comparison between mean and Softmin, we also compare the effect of Soft Sentinels.

\hypertarget{other-details}{%
\subsection{Other Details}\label{ssec:other-details}}

\hypertarget{adaptors-of-position-and-memory}{%
\subsubsection{Adaptors of Position and Memory}\label{sssec:adaptors-of-position-and-memory}}

For the sake of flexibility and performance, we usually do not use
inputted features \(x\) directly, but to first convert \(x\) into
\emph{position} \(f(x)\). Besides, the
training labels are not always consistent with the models' output. For
example, sometimes labels stand for discrete possibilities while the model outputs
logits. To adapt to this situation, we generally wrap the memory with an
adaptor function \(m\) such as a one-layer MLP, getting
\[\hat y | y_0 = m(y_0) + (f(x) - f(x_0)) \cdot g(x)\]
where \(g(x)\) stands for \emph{gradient}.

\hypertarget{the-choice-of-references}{%
\subsubsection{The Choice of References}\label{sssec:the-choice-of-references}}

We investigate four modes to choose references:

\begin{enumerate}
\item
  \textbf{Fixed}. When the inputs are rich-featured (different inputs
  are distinguishable simply by features), we can precompute the feature
  distances among data and find \(K\) nearest neighbours for each input
  as its fixed references.
\item
  \textbf{Random}. References are sampled arbitrarily.
\item
  \textbf{Dynamic NN}. \(K\) nearest neighbours according to the
  distance of position \(f(x)\) (not \(x\) as in Fixed Mode) are
  selected to be references. Since \(f(x)\) are dynamically changed
  following the updating of \(f\), this mode may require heavy
  computations.
\end{enumerate}

When the encodings of nodes are dynamic and no features
are provided, we usually adopt Random Mode in the training phase
for efficiency and adopt Dynamic NN Mode in the evaluation phase
for performance.

In experiments of Section~\ref{ssec:on-hyperparameters} that we carry on
dataset ogbl-ddi, we record both the scores in Random Mode and Dynamic NN
Mode in the evaluating phase.

\hypertarget{multiple-heads}{%
\subsubsection{Multiple Heads}\label{sssec:multiple-heads}}

Multi-heads can be a solution when it is
hard to boost performance by adding more parameters in a single
structure. It applies a model in separate instances. Each instance has
the potential to learn embeddings from different subspaces. So it can
also be regarded as an approach of Sampling.

We implement multi-heads in MAD Learning by combining the results from
separate instances with mean function.

\hypertarget{binary-mad-learning}{%
\subsection{Binary MAD Learning}\label{ssec:binary-mad-learning}}

\hypertarget{link-prediction}{%
\subsubsection{Link Prediction}\label{sssec:link-prediction}}

We model the relationship between a pair of nodes in a
graph by extending MAD Learning into binary situations.

Like what we do in the previous section, we first assume that the
relation \(r\) between node \(u\) and node \(v\) is a differentiable
function: \(r = r(u, v)\).
And besides the to-predict pair \((u, v)\), we also have another pair of
nodes \((u_0, v_0)\) called a \emph{reference}, with their relation
\(r_0\) already known.

We apply Total Derivative and Mean Value Theorem for Definite Integrals,
getting:

\[\begin{aligned}
r &= r(u, v) \\
&= r_0 + \int_{(u_0, v_0)}^{(u, v)} dr \\
&= r_0 + \int_{(u_0, v_0)}^{(u, v)} (\frac{\partial{r}}{\partial{u}} du + \frac{\partial{r}}{\partial{v}} dv) \\
&\approx r_0 + (u - u_0) \cdot \frac{\partial{r}}{\partial{u}}|_{(u, v)} + (v - v_0) \cdot \frac{\partial{r}}{\partial{v}}|_{(u, v)} \\
&\triangleq \hat r | r_0
\end{aligned}\]

To simplify the above model and assure the reference \((u_0, v_0)\)
as close to the to-predict pair \((u, v)\) as possible, we set
\(u_0 = u\) or \(v_0 = v\), meaning \((u, v)\) always shares with a
common node with \(r(u_0, v_0)\).

When \(v = v_0\) holds, the partial differential
\(\frac{\partial{r}}{\partial{u}}|_{(u, v)}\) can be regarded as the
change of \(r(u, v)\) after slightly moving the node \(u\) to \(u_0\)
but with node \(v\) fixed, as depicted in Figure~\ref{fig:mad-flow}(b).

Therefore, we may further assume
\(\frac{\partial{r}}{\partial{u}}|_ {(u, v)} = g_1(v)\) if \(v = v_0\)
and \(\frac{\partial{r}}{\partial{v}}|_ {(u, v)} = g_2(u)\) if
\(u = u_0\), making \[\begin{cases}
\hat r|r_0 = g_1(v) \cdot (u - u_0) + r_0, &\text{if } v = v_0 \\
\hat r|r_0 = g_2(u) \cdot (v - v_0) + r_0, &\text{if } u = u_0
\end{cases}\]
Here \(g_1(\cdot)\) is \emph{destination differential function}
and \(g_2(\cdot)\) is \emph{source differential function}.
If the edge is undirected, these two functions can be shared.

A direct application in this binary situation is to predict
whether a pair of nodes in a graph are likely to be connected.
We test our method on dataset ogbl-ddi from OGB and MAD Learning
outperforms SotA baselines by a large margin but with fewer parameters.

\hypertarget{the-geometric-meaning-of-mad-learning}{%
\subsubsection{The Geometric Meaning of MAD Learning}\label{sssec:the-geometric-meaning-of-mad-learning}}

\begin{figure}
\centering
\includegraphics[bb=0 0 800 400, width=0.66\textwidth]{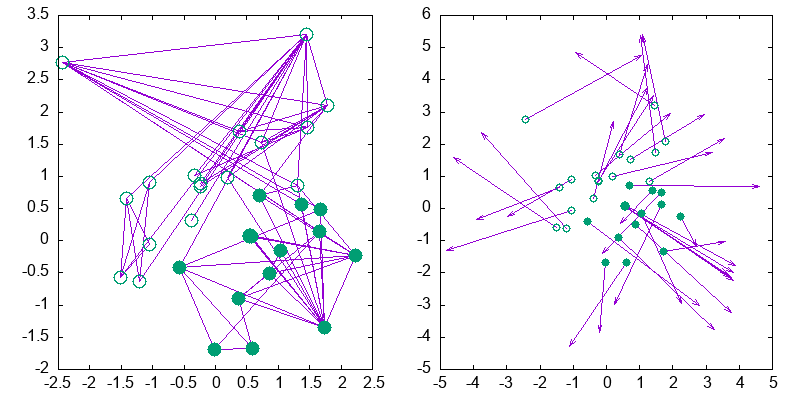}
\caption{
  2-dimensional encodings learnt from connections in Zachary's Karate
  Club. Nodes are placed with their 2-dimensional positions as coordinates
  and coloured according to their groups. In the left plot, nodes are
  connected if corresponding club members interact. In the right plot,
  2-dimensional gradients are visualized as vectors attached to nodes.
}
\label{fig:karate}
\end{figure}

To reveal the meaning of the positions and gradients within MAD
Learning, in Figure~\ref{fig:karate}, we visualize the 2-dimensional
encodings learnt from only connections in Zachary's Karate Club,
a social network representing the interaction among club members from
two communities. In both plots, each member as a node is placed at
her 2-dimensional position and coloured according to which
group she belongs.

In the left plot, the nodes are positioned geometrically into two clusters.
This implies that the positions may have enough information for downstream tasks
such as Node Classification~\cite{1403.6652v2}. And as we discussed
earlier when comparing with GCN in Section ~\ref{sec:related-works}, connected
nodes have not to be close to each other, while neighbours of the same
node tend to stay in nearby positions.

In the right plot, we see each node is attached with its gradient as a
vector, pointing regularly opposite to the centre. More precisely, a
node's gradient is the direction along with which other nodes become
more and more `connectable'.

\hypertarget{experiments}{%
\section{Experiments}\label{sec:experiments}}

If not mentioned, the following experiments~\footnote{
    Code: \url{https://github.com/cf020031308/mad-learning}
} in this work use Adam~\cite{1412.6980v9}
as optimizers with their learning rate set to 0.005, set K = 8, run in Random Mode
in the training phase and in Dynamic NN Mode in the evaluating phase, encode positions
and gradients into 32-dimensional vectors, disable multi-heads, and mix Softmin with
8 Soft Sentinels distant at 1.

\hypertarget{on-hyperparameters}{%
\subsection{On Hyperparameters}\label{ssec:on-hyperparameters}}

In this section, we evaluate our method on the dataset \emph{ogbl-ddi}
from Open Graph Benchmark (OGB) to analyse hyperparameters
of MAD Learning. The metric is \emph{Hits@20}, the rate of true connections
that are ranked higher than the 20 top-ranked but false ones.

In the training phase, we sample arbitrary pairs of nodes to construct
negative samples~\cite{1607.00653v1} and compare the scores between
connected pairs and negative samples with Cross-Entropy as the loss function:
\[L = - \sum\limits_{i=1}^{y} \log (p_y(i)) - \sum\limits_{i=1}^{n} \log(1 - p_n(i))\]
where \(y\) is the number of positive samples and \(n\) of negative samples,
\(p_y(i)\) is the predicted probability of the \(i\)-th positive sample and \(p_n(i)\) 
of the \(i\)-th negative sample.

In the evaluating phase, we record the scores not only in Dynamic NN Mode but also
in Random Mode.

\hypertarget{memory-and-differential-functions}{%
\subsubsection{Memory and Differential Functions}\label{sssec:memory-and-differential-functions}}

To measure how important the memory and the differential functions are,
we experiment with different parts of MAD Learning disabled:

\begin{enumerate}
\item
  \textbf{mad}. Complete MAD Learning.
\item
  \textbf{nograd}. MAD with gradients \( g_1(v) = g_2(u) = 0 \).
  Only memory and the Softmin weights are involved.
\item
  \textbf{nomem}. MAD with memory \( r_0 \) = 0.
\end{enumerate}

\begin{figure*}
\centering
\includegraphics[bb=0 0 1000 375, width=1.33\textwidth]{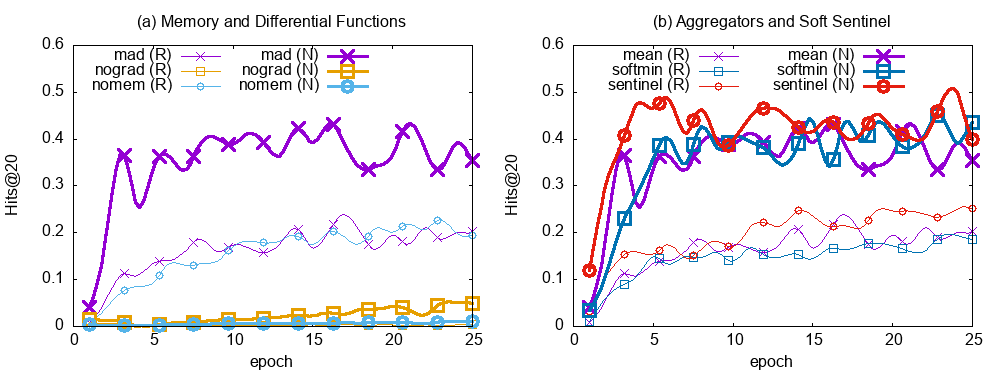}
\caption{
  Hits@20 on ogbl-ddi in 25 epochs.
  \textbf{R} in brackets stands for Random Mode in the evaluating phase
  and \textbf{N} for Dynamic NN Mode
}
\label{fig:hyperparams}
\end{figure*}

The results are depicted in Figure~\ref{fig:hyperparams}(a). The performance of the
complete MAD Learning in Dynamic NN Mode significantly surpasses all others,
proving that both memory and inference are indispensable.

Besides, we notice that in Random Mode, having no memory still works,
because in such way the referenced nodes \(u_0\) and \(v_0\) can be regarded as some other
pseudo nodes located at the same positions but with 0 logits to connect to \(v\) or \(u\).

Furthermore, Matrix Factorization can be reduced to MAD Learning
with no memory but only one fixed reference of \(v\) and a pseudo node located at
the origin point, as \[\hat r = g_1(v) \cdot (u - 0) + 0 = g_1(v) \cdot u\]

Another discovery is that the performance without memory in Dynamic NN Mode
is far worth than in Random Mode. This is because references in Dynamic
NN Mode are too close to contribute enough differences to reach
the scale of predictions without memory.

\hypertarget{aggregators-and-soft-sentinel}{%
\subsubsection{Aggregators and Soft Sentinel}\label{sssec:aggregators-and-soft-sentinel}}

We have these three experimental settings to examine the contribution of
Softmin and Soft Sentinels:

\begin{enumerate}
\item
  \textbf{mean}. Estimations are aggregated by mean function.
\item
  \textbf{softmin}. Estimations given by different references are summed
  up weighted by the results of Softmin applied to the distances.
\item
  \textbf{sentinel}. Estimations of softmin with 8 Soft Sentinels at distance 1 added.
\end{enumerate}

As is shown in Figure~\ref{fig:hyperparams}(b), it is no much difference between
mean and Softmin. But when mixed with Soft Sentinels, MAD Learning performs
better and converges faster.

\hypertarget{image-recognition}{%
\subsection{Image Recognition}\label{ssec:image-recognition}}

We conduct experiments on Image Recognition as an application of unary MAD Learning.

The datasets we use are MNIST~\cite{lecun1998gradient}, KMNIST~\cite{clanuwat2018deep},
CIFAR-10 and CIFAR-100~\cite{krizhevsky2009learning}. The
baselines are a two-layered convolutional neural network notated as
\emph{ConvNet} and ResNet18~\cite{1512.03385v1}.
In MAD Learning we separately use the above ConvNet or ResNet18 to
extract image features before mapping them into positions and gradients.
The two variances of MAD Learning with different features extractors are
notated as \emph{MAD-conv} and \emph{MAD-18}.

\begin{table}
\centering
\begin{tabular}{llrr}
\toprule
Data Name & Method & \#Params & Accuracy\\
\midrule
MNIST      & ConvNet           & 60074    & 98\%\\
           & \textbf{MAD-conv} & 53982    & 98\%\\
KMNIST     & ConvNet           & 60074    & 94\%\\
           & \textbf{MAD-conv} & 53982    & 94\%\\
CIFAR-10   & ResNet18          & 11181642 & 82\%\\
           & \textbf{MAD-18}   & 11244338 & 82\%\\
CIFAR-100  & ResNet18          & 11227812 & 53\%\\
           & \textbf{MAD-18}   & 11808368 & 50\%\\
\bottomrule
\end{tabular}
\caption{Image Recognition}
\label{tbl:image-recognition}
\end{table}

\begin{table*}\centering
\begin{tabular}{lrr}
\toprule
 Method & Hits@20 & \#Params\\
\midrule
\textbf{MAD} & \textbf{0.6781} & \textbf{1228897}\\
LRGA + GCN   & 0.6230          & 1576081\\
GCN + JKNet  & 0.6056          & 1421571\\
GraphSAGE    & 0.5390          & 1421057\\
GCN          & 0.3707          & 1289985\\
\bottomrule
\end{tabular}
\caption{Link Prediction on ogbl-ddi}
\label{tbl:ogbl-ddi}
\end{table*}

We train these models for 50 epochs and record their best accuracy
scores every 5 epochs, which are summarized in
Table~\ref{tbl:image-recognition}. As we
can see, MAD Learning has no advantage in this application.
We suggest that MAD Learning
is better at complex tasks involving both memory and inference.
Since Image Recognition is a intuitive task as ``You know it
when you see it'', MAD Learning can do no better than convolutional
networks.

However, we repeat that MAD Learning does not predict directly.
From another point of view, this experiment implies that undirect
references can also be beneficial on par with direct information.

\hypertarget{link-prediction-1}{%
\subsection{Link Prediction}\label{ssec:link-prediction}}

%
%

We compare the performance of MAD Learning, implemented
with 12 heads, 12-dimensional positions and 12-dimensional gradients,
against SotA baselines from the top of the leaderboard on OGB, including GCN,
GraphSAGE~\cite{hamilton2017inductive}, JKNet~\cite{xu2018representation},
and LRGA~\cite{2006.07846v1}.

Results in Table~\ref{tbl:ogbl-ddi} show that MAD Learning can achieve a
higher Hits@20 score with fewer parameters, thus producing the new SotA.

\hypertarget{discussion}{%
\section{Discussion}\label{sec:discussion}}

Most experiments in this work are conducted on predicting links where
the relation \(r\) represents a logit, but it is not difficult to explain
it as edge weights.
And by extending it from a scalar to a vector, MAD Learning can be used
for graphs with featured edges.

We also point out that MAD Learning can learn relations in heterogeneous
graphs where nodes belong to different types (usually represented by
encodings in different lengths). The only requirement is that positions
of the source nodes should match with gradients of the destination nodes
and vice versa.

For example, in Recommender Systems, we can encode positions of users
and gradients of items in 8-dimensional vectors, and encode positions of
items and gradients of users in vectors with different dimensions, say
16.

For ternary relations such as head-relation-tail triplets \(f=f(h, r, t)\)
in Knowledge Graphs, we may also extend the binary MAD Learning into:
\[\begin{cases}
\hat f|f_0 = g_1(h, r) \cdot (t - t_0) + f_0, &\text{if }h = h_0, r = r_0\\
\hat f|f_0 = g_2(r, t) \cdot (h - h_0) + f_0, &\text{if }r = r_0, t = t_0\\
\hat f|f_0 = g_3(t, h) \cdot (r - r_0) + f_0, &\text{if }t = t_0, h = h_0
\end{cases}\]
The same extension can be made in multi situations.

\hypertarget{conclusion}{%
\section{Conclusion}\label{sec:conclusion}}

In this work, we explore a novel learning paradigm which is flexible,
effective and interpretable.
The outstanding results, especially on Link Prediction, open the door
for several research directions:

\begin{enumerate}
\item
  The most important part of MAD Learning is memory. However, MAD
  Learning have to index the whole training data for random access. In
  Link Prediction, we implement memory as a dense adjacency matrix which
  results in huge occupation of space. The way to shrink memory and
  improve the utilization of space should be investigated in the future.
\item
Based on memory as the ground-truth, MAD Learning appends some difference as the second part. We implement this difference simply as the product of distance and differential function, but we believe there exist different ways to model it.
\item
  The third part of MAD Learning is the similarity, which is used to assign weights to estimations given by different references. We reuse distance to compute the similarity, but decoupling it by some other embeddings and some other measurements such as inner product should also be worthy to explore.
\item
  In this work, we do deliberately not combine direct information to
  focus only on MAD Learning. Since MAD Learning takes another parallel
  route to predict, we believe integrating MAD Learning and Conventional
  Supervised Learning is also a promising direction.
\end{enumerate}

\bibliographystyle{named}
\bibliography{mad}

\end{document}